\documentclass[times,twocolumn,final,authoryear]{elsarticle}

\usepackage{ycviu}
\usepackage{framed,multirow}

\usepackage{amssymb}
\usepackage{latexsym}

\usepackage{url}
\usepackage{xcolor}
\usepackage{hyperref}

\definecolor{newcolor}{rgb}{.8,.349,.1}

\journal{Computer Vision and Image Understanding}

\begin{document}

\begin{frontmatter}

\title{Visual search over billions of aerial and satellite images}

\author[1]{Ryan \snm{Keisler}\corref{cor1}} 
\ead{ryan@descarteslabs.com}
\author[1]{Samuel W. \snm{Skillman}}
\author[1]{Sunny \snm{Gonnabathula}}
\author[1]{Justin \snm{Poehnelt}}
\author[1]{Xander \snm{Rudelis}}
\author[1]{Michael S. \snm{Warren}}

\address[1]{Descartes Labs, 1613 Paseo De Peralta, Santa Fe, NM, 87501, USA}

\received{}
\finalform{}
\accepted{}
\availableonline{}
\communicated{r}

\begin{abstract}
We present a system for performing visual search over billions of aerial and satellite images.  The purpose of visual search is to find images that are visually similar to a query image.  We define visual similarity using 512 abstract visual features generated by a convolutional neural network that has been trained on aerial and satellite imagery.  The features are converted to binary values to reduce data and compute requirements.  We employ a hash-based search using \texttt{Bigtable}, a scalable database service from Google Cloud.  Searching the continental United States at 1-meter pixel resolution, corresponding to approximately 2 billion images, takes approximately 0.1 seconds.  This system enables real-time visual search over the surface of the earth, and an interactive demo is available at \url{https://search.descarteslabs.com}.
\end{abstract}


\end{frontmatter}


\section{Introduction}

Visual search, the ability to find images that are visually similar to a query image, is a form of pattern recognition applied to the problem of image retrieval.  An image retrieval system allows the user to find images that have some logical connection to a set of query parameters such as keywords, captions, or the in the case of content-based image retrieval (CBIR), the visual content of the image itself \citep{gudivada1995content, rui1999image, smeulders2000content, liu2007survey, datta2008image}.  CBIR systems use computer vision techniques (e.g.\ the SIFT descriptor, \cite{lowe2004distinctive}) to quantify texture, color, and shapes present in an image and thereby define image similarity.  Convolutional neural networks \citep{lecun1995convolutional} trained on large image datasets \citep{deng2009imagenet} provide another means to extract rich feature representations of photographic imagery, and can be used in CBIR systems for visual search \citep{sharif2014cnn, babenko2014neural, yue2015exploiting, tolias2015particular, zheng2018sift}.

Visual search is an increasingly common offering from large software companies.  Some of these systems are focused on consumer products, e.g. Pinterest Lens \citep{jing2015visual, zhai2017visual}, while other search systems from Google, Microsoft, Baidu, or Yandex operate on photographic imagery more generally.

Aerial and satellite imagery provides another large dataset on which to build a visual search system, although there is much less prior work relative to visual search over photographic imagery.  A significant milestone came in May 2016, when a team from Carnegie Mellon University released Terrapattern\footnote{\url{http://www.terrapattern.com}}, an interface for visual search over seven metropolitan areas.  The work presented in this paper was largely inspired by Terrapattern.  We sought to scale a visual search system beyond metropolitan areas to an entire country or the whole world.


Going beyond searching a single imagery layer, it is interesting to consider what visual search over all publicly available aerial and satellite imagery would look like.  As organizations like Descartes Labs\footnote{\url{https://descarteslabs.com}}, Google Earth Engine\footnote{\url{https://earthengine.google.com}}, and Sentinel Hub\footnote{\url{https://www.sentinel-hub.com/}} continue to build systems to ingest this flood of remote sensing data, what information do they store and in what format?

To begin with, there is the image data itself, which is currently at the petabyte scale and will  cross into tens of petabytes in the coming decade.  Beyond that, metadata such as the geographical extent of the image, the time at which the image was acquired, and the sensor used to acquire the image are clearly valuable; one can construct databases to efficiently index and search over these metadata.  One can also store metadata related to the image data itself: basic summary statistics or the fraction of pixels covered in clouds.  What is lacking is metadata encoding a deeper visual or semantic understanding of the image content.  Ideally, this ``visual metadata'' would be significantly smaller than the original image data, say at least 100X smaller, and, like the spatial and temporal metadata described above, would encode some meaningful notion of ``distance'' between images.  The feature vectors presented in this paper provide a step in that direction.

How will such visual metadata be used for real applications?  Possibilities include:

\begin{itemize}
    \item interactive exploration of the earth's surface: a user wants to quickly search for visually similar objects, and that is the end goal.
    \item ground truth: a user could efficiently gather ground truth to be used to train a downstream computer vision model.
    \item pre-processing filter: a user needs to run a computer vision model over a large region but, in order to reduce computational expense, would like to limit the processing to only those images that might contain the object of interest.
\end{itemize}

This paper is organized as follows.  In Section~\ref{sec:imagery} we describe the two image datasets used in this work.  In Section~\ref{sec:features} we describe how we modified a convolutional neural network to extract useful binary features from this imagery, and in Section~\ref{sec:search} we detail the two search methods used to search over these features.  Finally, we show example search results in Section~\ref{sec:results} and conclude with a summary of future research directions in Section~\ref{sec:conclusion}.

\section{Imagery}
\label{sec:imagery}
We used two imagery sources in the work presented here:
\begin{itemize}
  \item \textbf{Aerial over USA - }We use aerial imagery from the National Agriculture Imagery Program\footnote{\url{https://lta.cr.usgs.gov/NAIP}} (NAIP) and the Texas Orthoimagery Program\footnote{\url{https://tnris.org/2015-statewide-orthoimagery-project/}}.  These datasets provide coverage of the lower 48 states of the United States.
  \item \textbf{Landsat 8 over Earth - }We created a custom, pan-sharpened, 15-meter global composite \citep{warren2016data} using data from Landsat 8, one of the workhorses of NASA’s satellite-based earth observation program.
\end{itemize}

We built a visual search system for each of these datasets, but the systems differ in how the features were generated, as described in Section \ref{sec:features}.


We maintain the projection of both datasets in the 60 Universal Transverse Mercator (UTM) projections.  These conformal projections preserve angles and have weak area distortion within a UTM zone.  Both of these properties are desirable for visual search, as we would only be making the search problem more difficult if we used a projection that introduced strong distortion of angles or areas.

We chunked this imagery into square image tiles, 128 pixels on a side.  Adjacent tiles overlap by 64 pixels in the vertical and horizontal directions.  While this overlapping tiling scheme requires two times more data than a non-overlapping scheme, it gives any object the chance to be roughly centered in at least one image tile.

We used the Descartes Labs Platform\footnote{\url{http://docs.descarteslabs.com}} to  access and process this imagery.  This platform allows the user to search for the existence of imagery, quickly load that imagery into memory in the form of a python \texttt{ndarray}, and parallelize computations over tens of thousands of CPUs.

\section{Feature Vectors}
\label{sec:features}

We start with a convolutional neural network with a 50-layer ResNet architecture \citep{he2016deep}, pre-trained on ImageNet\footnote{\url{http://www.image-net.org}}.  This network is conveniently provided pre-trained in the Keras\footnote{\url{https://keras.io}} deep learning package, which we used with a Tensorflow\footnote{\url{https://www.tensorflow.org/}} back-end.

In our initial experiments we used the ``out of the box'' features generated in the last few layers of the ImageNet-trained network.  These layers give surprisingly good similarity search results for satellite imagery, despite being trained on photographic imagery of animals, plants, vehicles, etc.  But in the end we modified the network in two ways, as described below.

\begin{figure*}[!ht]
\centering
\includegraphics[scale=0.7]{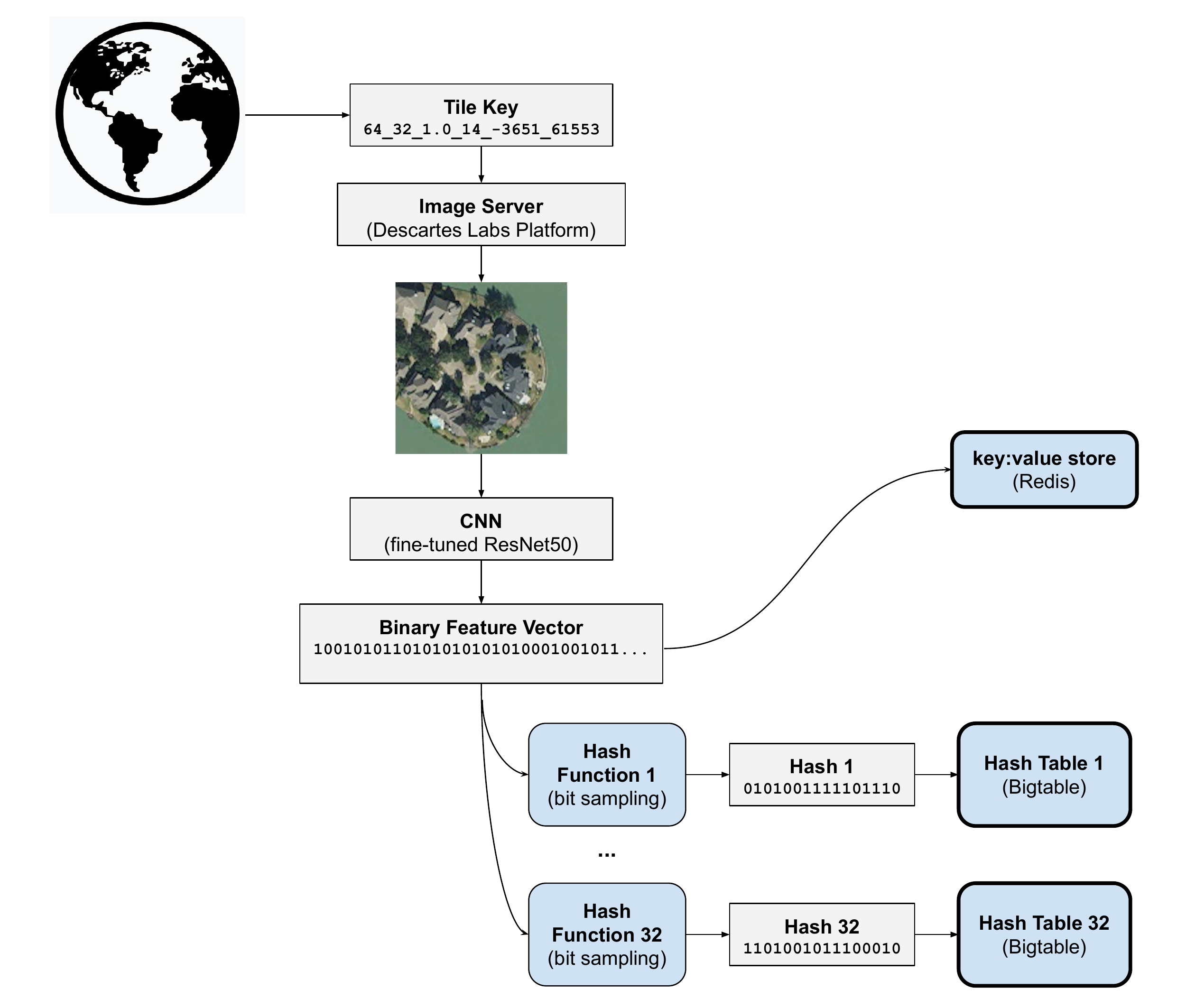}
\caption{The feature-generation architecture.  The earth is tiled into a set of tiles, and aerial or satellite imagery is retrieved for each tile using the Descartes Labs Platform.  Each image is passed through a convolutional neural network (CNN) that has been fine-tuned for aerial or satellite imagery.  The resulting feature vector is binarized and stored in a simple key:value store, as well as a set of 32 hash tables used for approximate nearest neighbor search.}
\label{fig:featgen}
\end{figure*}

An overview of the feature-generation process is shown in Figure~\ref{fig:featgen}.  

\subsection{Binary Features}
We decided that we ultimately wanted to search over binary features, due to their smaller data footprint.  To that end, we encouraged the network to make floating-point features very close to 0.0 or 1.0 at the layer of interest by injecting noise with an amplitude comparable to the width of the layer's activation function \citep{salakhutdinov2009semantic}. The noise is injected during training but not during inference.  The network learns to make almost-binary features at this layer; otherwise the noise destroys the information that the layer is trying to pass on. Finally, we binarize the floating-point features by thresholding at 0.5.

\subsection{Customizing for Aerial and Satellite Imagery}
We customized this network to work with each source of aerial and satellite imagery.  For NAIP, we followed the strategy used by Terrapattern and fine-tuned the network to classify images into 130 object classes from OpenStreetMap (OSM), such as parking lots or golf courses \citep{zheng2016good}.  Prior to fine-tuning, we added two fully-connected layers to the end of the network, and extracted 512 binary features from one of them \citep{vo2019generalization}.  We emphasize that this supervised learning step is not used to produce a network that is good at identifying particular object classes (although it does ultimately do better on these classes), but rather to create a network that produces features that are useful for generic visual search on aerial and satellite imagery.

For Landsat 8, the OSM classes were less useful due to the much coarser spatial resolution, so we took an unsupervised approach instead.  Specifically we used an autoencoder to compress the 2048 features from the penultimate layer of ResNet50 into 512 binary features using a 2048-1024-512-1024-2048 autoencoder architecture.  The autoencoder was trained by passing Landsat 8 imagery through the original ResNet50 network.

We also note that we do not directly optimize on the binary features used for look-up, unlike many feature hashing approaches \citep{norouzi2012hamming}. Instead, we rely on the fact that images of similar classes do map to similar outputs of late layers in the neural net.

At the end of this process, we have mapped the original 128x128x3 8-bit image (or 16-bit image in the case of Landsat 8) to 512 bits.  This corresponds to a 768X (1536X) reduction in data size.  These binary features form a compact representation of the visual information present in each image.

We pre-compute the feature vectors for all of the tiles in each dataset: approximately 2 billion tiles for NAIP and 200 million tiles for Landsat 8. We distributed this computation across tens of thousands of CPUs in the Google Cloud Platform.

We store the binary features vectors in \texttt{Redis}, an in-memory data store.  The data is stored in key:value pairs.  The keys are strings that uniquely identify each image tile, and the values are the corresponding 512-bit feature vectors, each stored as 64 bytes.

\section{Search}
\label{sec:search}

Now that we have feature vectors, how do we search for similar vectors?

We first define a distance between vectors.  We use the Hamming distance: the number of bits that differ between the two binary vectors.  A small distance is associated with visual similarity.

Next we need to find the $k$-nearest vectors to a query vector. We use two methods: an exact, brute-force search, and an approximate, hash-based search, as described below.

\subsection{Direct Search}
By taking advantage of low-level instructions for comparing bits (\texttt{XOR}) and counting bits (\texttt{\_\_builtin\_popcountll}), we can perform a direct, exact, brute-force search over the 200 million Landsat 8 images.  Search for similar vectors is a nearest-neighbors search in the 512-bit space.  This works as follows.

\begin{itemize}
    \item A query image tile is selected.
    \item We retrieve the binary feature vector for this image tile using \texttt{Redis}.
    \item We compute the Hamming distance between the query vector and all vectors using a pre-computed binary file that contains all binary feature vectors.  This sub-program is written in C, using the GCC compiler with \texttt{-Ofast} optimization.
    \item We keep the $k$-nearest feature vectors and determine their corresponding images tiles using a second index file.
\end{itemize}
This direct, brute-force method searches across the 200 million Landsat 8 images in approximately 2 seconds using a single thread.  While an individual query is single-threaded, we run a server (16 virtual CPUs, Haswell architecture, 32 GB of RAM) which allows for up to 16 concurrent queries that share the feature vector data in memory.

\subsection{Hash-based Search}
The NAIP dataset is approximately ten times larger than the Landsat 8 dataset, with approximately 2 billion images.  While the direct search method works, it is too slow for interactive use.  Instead we use an approximate, hash-based search method.  The general idea is to use hash functions to quickly return a list of candidate neighbors, and then run a brute-force search on that relatively small list of candidates.

\begin{figure*}[!ht]
\centering
\includegraphics[scale=0.7]{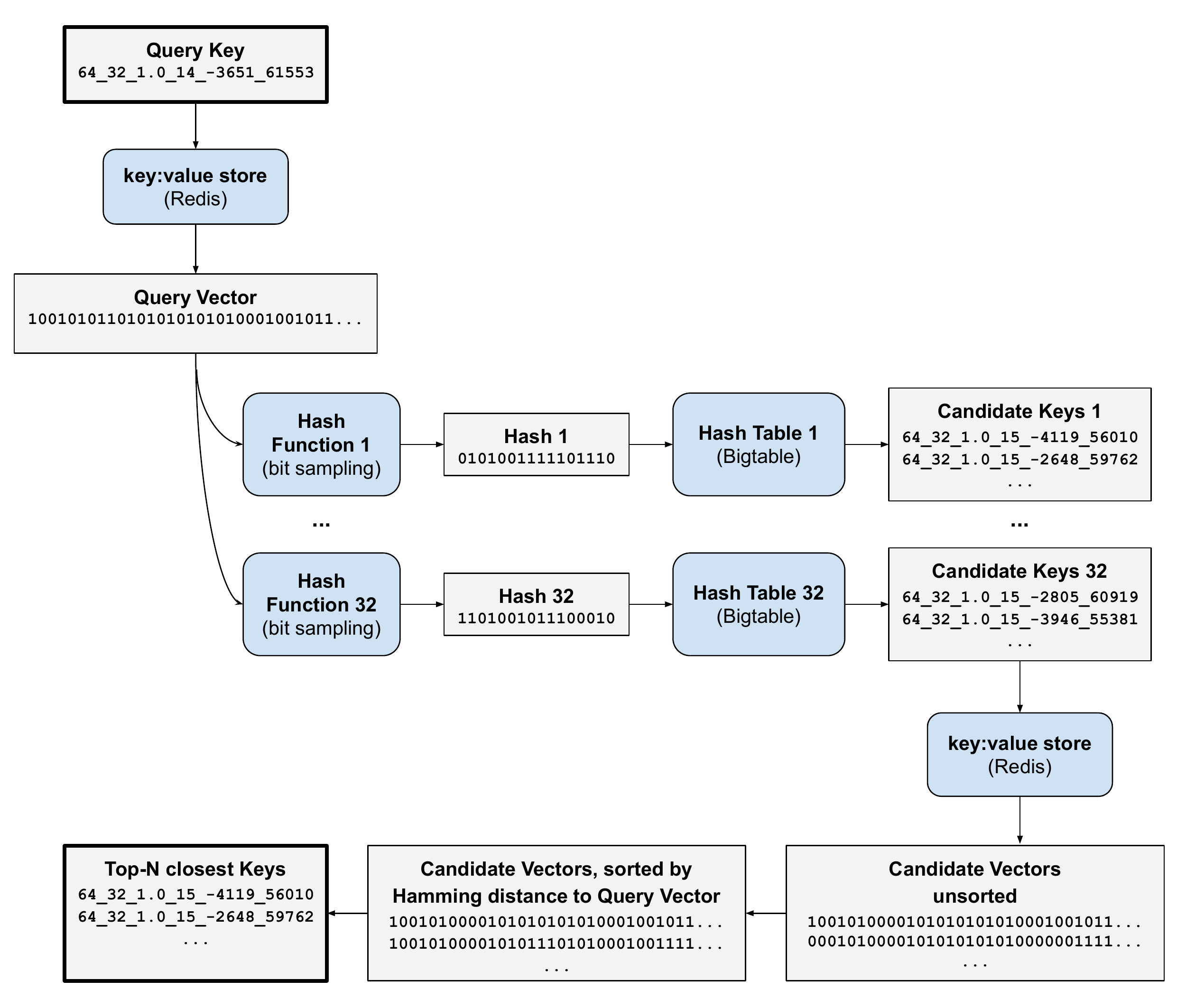}
\caption{The hash-based search architecture.  A query image is selected by the user, and the binary feature vector for this image is retrieved from a key:value store.  After hashing this feature vector through a set of 32 hash tables, we retrieve a set of candidate neighbors.  Finally, these candidates are sorted by distance to the original query vector, and the top-$N$ image keys are returned.}
\label{fig:hashsearch}
\end{figure*}

More specifically, we use bit sampling, a simple form of locality-sensitive hashing.  We use a family of 32 hash functions, each of which simply selects a subset of 16 bits from the full, 512-bit feature vector.  Within each hash table, the keys are specific realizations of those 16 bits, e.g. 0101001111101110, and the values are lists of strings (image tile ids) corresponding to tiles that contain those specific realizations of those 16 bits.  We store the hash tables in \texttt{Bigtable}, a scalable NoSQL database service from Google Cloud.  An overview of the hash-based search system is shown in Figure~\ref{fig:hashsearch} and works as follows.

\begin{itemize}
    \item A query image tile is selected.
    \item We retrieve the binary feature vector for this image tile using \texttt{Redis}.
    \item We apply each of the 32 hash functions to this feature vector, each of which returns a set of candidate neighbors.
    \item We compute the unique set of candidate neighbors and fetch their corresponding feature vectors using \texttt{Bigtable}.
    \item We perform a brute-force distance calculation between the query vector and the candidate vectors.
    \item We keep the $k$-nearest feature vectors and their corresponding images tile ids.
\end{itemize}
The search takes approximately 0.1 seconds per query over approximately 2 billion images.


\section{Results}
\label{sec:results}

\subsection{Search Results}
\begin{figure}[!ht]
\centering
\includegraphics[width=\columnwidth]{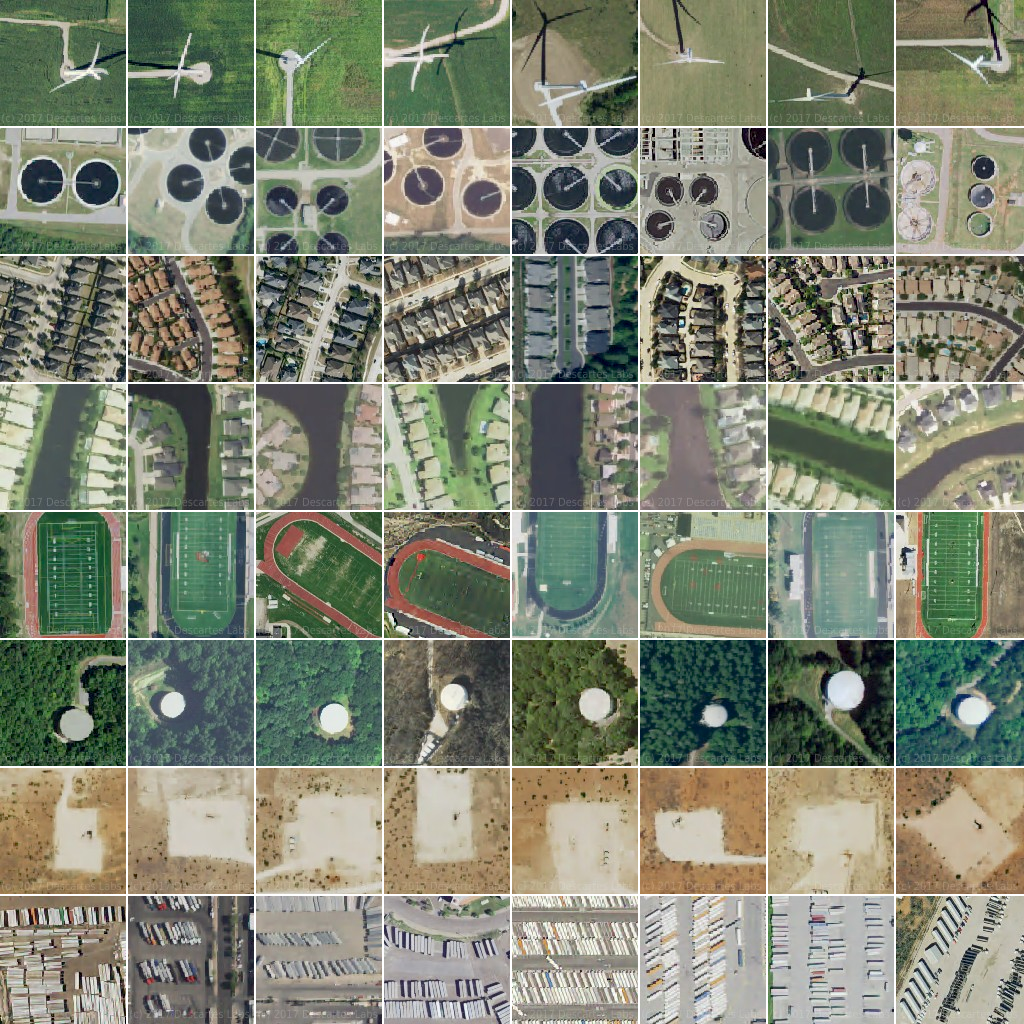}
\caption{Example search results of man-made features from NAIP.  The leftmost image of each row is the query image, and the rest of the row shows the top seven results.  These queries show wind turbines, water treatment plants, suburban homes, waterfront homes, stadiums, isolated storage tanks, oil and gas sites, and containers.}
\label{fig:manmade}
\end{figure}

\begin{figure}[!ht]
\centering
\includegraphics[width=\columnwidth]{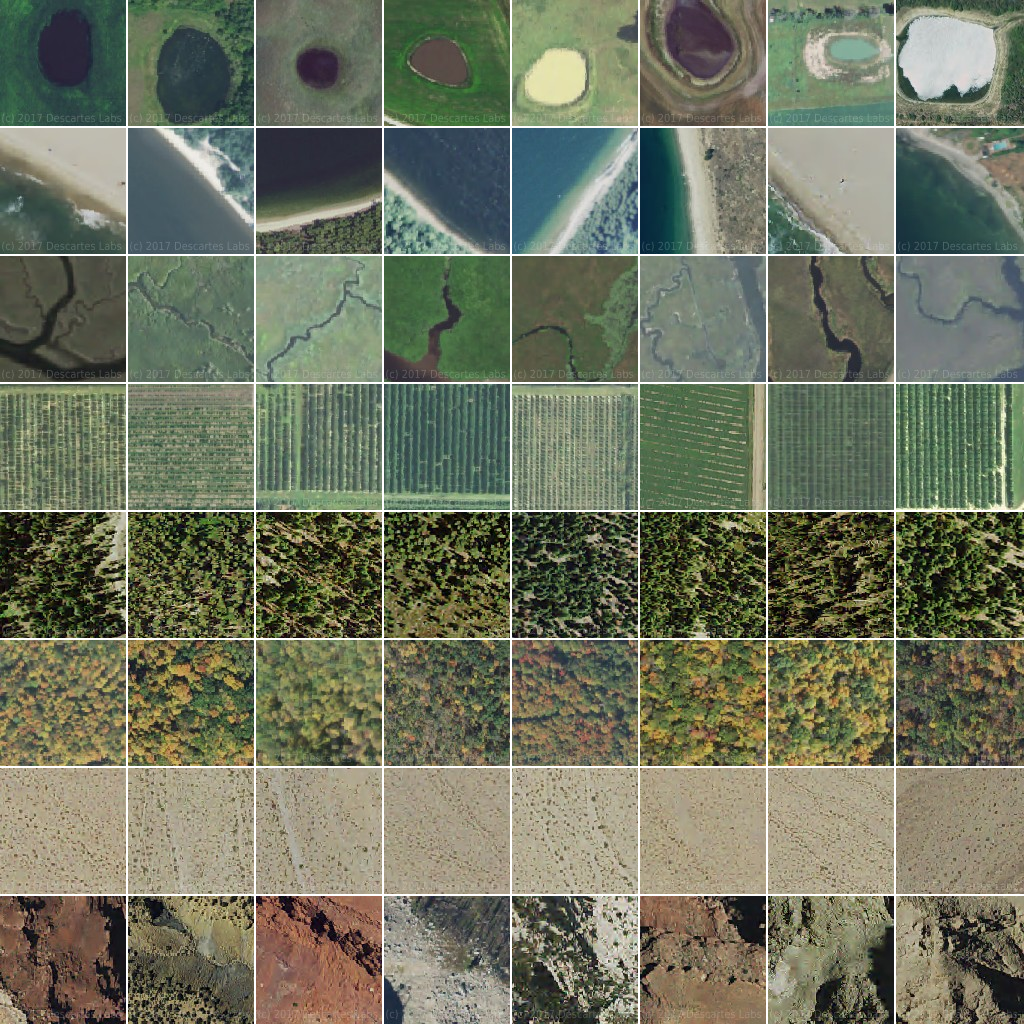}
\caption{Example search results of natural features from NAIP.  The leftmost image of each row is the query image, and the rest of the row shows the top seven results.  These queries show ponds, beaches, sinuous rivers, orchards, forest, trees changing color, desert, and rocky outcroppings.}
\label{fig:natural}
\end{figure}

Search results from NAIP are shown in Figures~\ref{fig:manmade} and \ref{fig:natural}.  These search queries span a wide range of object types and landscapes, from industrial infrastructure to subtle natural features.

We expect to have relatively good search results for many of the queries shown here (e.g. wind turbines) because the query object was one of the 130 OSM classes used to fine-tune the NAIP network.  However, the system also provides good search results over generic images that were not part of the supervised learning phase, such as the sinuous waterways or the trees changing color.

\begin{figure}[!ht]
\centering
\includegraphics[width=\columnwidth]{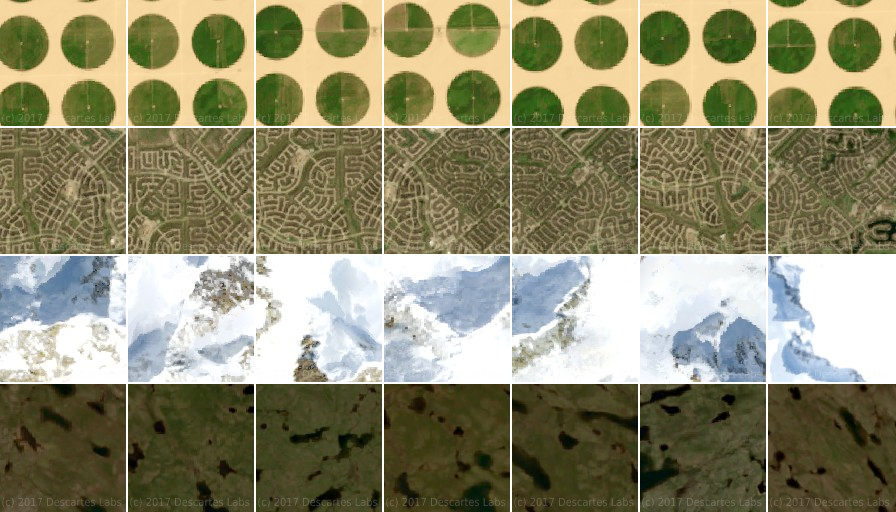}
\caption{Example search results from Landsat 8.  The leftmost image of each row is the query image, and the rest of the row shows the top seven results.  These queries show center-pivot irrigation, suburbs, snow-capped mountains, and arctic ponds.}
\label{fig:l8}
\end{figure}

Search results from Landsat 8 are shown in Figure~\ref{fig:l8}.  The Landsat 8 features were generated without any supervised learning or reference to OSM classes, so the search results demonstrate generic visual search.

\begin{figure}[!ht]
\centering
\includegraphics[width=\columnwidth]{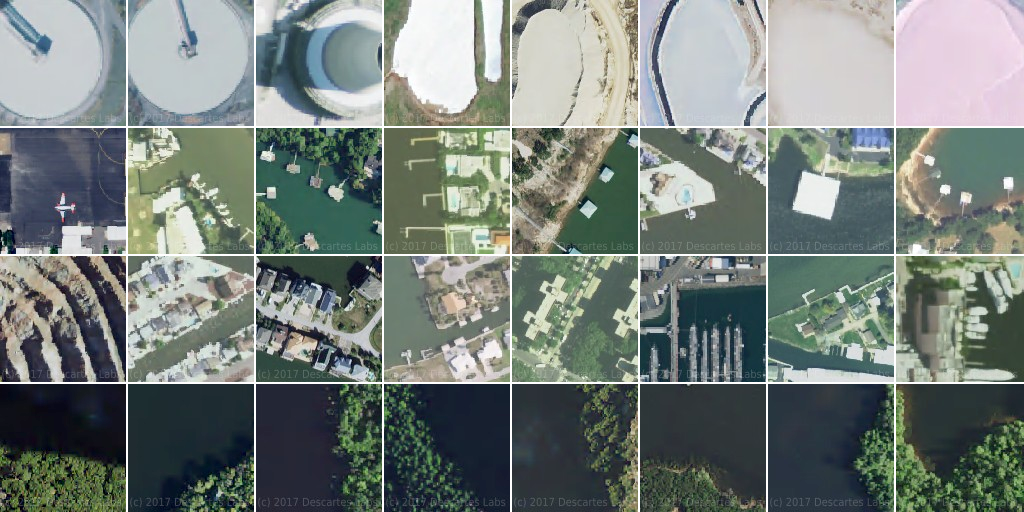}
\caption{Low-quality search results for a water treatment plant, a small airport, a mine, and a forest in shadow.  While these results do contain images that are visually similar to the query image, they are failures in the sense that they do not contain the query object class.}
\label{fig:bad}
\end{figure}

We show low-quality search results in Figure~\ref{fig:bad}.  While these results do contain images that are visually similar to the query image, they are failures in the sense that they do not contain the query object class.  This highlights the fact that there is a balance to strike between providing good search results for common object classes and providing good search results for generic, class-agnostic search.

Finally, we have made a quantitative assessment of search quality for 20 query types.  For each query we have manually recorded the precision (fraction of true positives) out of the top-30 search results.  We have not attempted to assess recall, given the lack of unified, comprehensive ground truth for these object types.

Our top-30 precision results are as follows.  For the 10 object types that were present during the network fine-tuning stage, the average top-30 precision is 86\%.  The individual results for this category are: airplanes (36\%), baseball diamonds (100\%), football fields (50\%), golf courses (100\%), highway overpasses (70\%), marinas (100\%), parking lots (100\%), railyards (100\%), storage tanks (100\%), and taxiways (100\%).

For the 10 object types that were not present during the fine-tuning stage, the average top-30 precision is 58\%.  The individual results for this category are: barges (10\%), CAFOs (97\%), center pivot irrigation (97\%), cul-de-sacs  (33\%), dry docks (3\%), feed lots (50\%), roller coasters (3\%), row housing (90\%), suburban homes (100\%), and shipping containers (93\%).  As one would expect, the search results are generally of higher quality for object types that were present during the network fine-tuning stage (86\% vs 58\%), although the precision of the new object types (58\%) is still high enough to be useful for some applications, e.g. efficiently gathering ground truth to be used to train a downstream computer vision model.

\begin{figure}[!ht]
\centering
\includegraphics[width=\columnwidth]{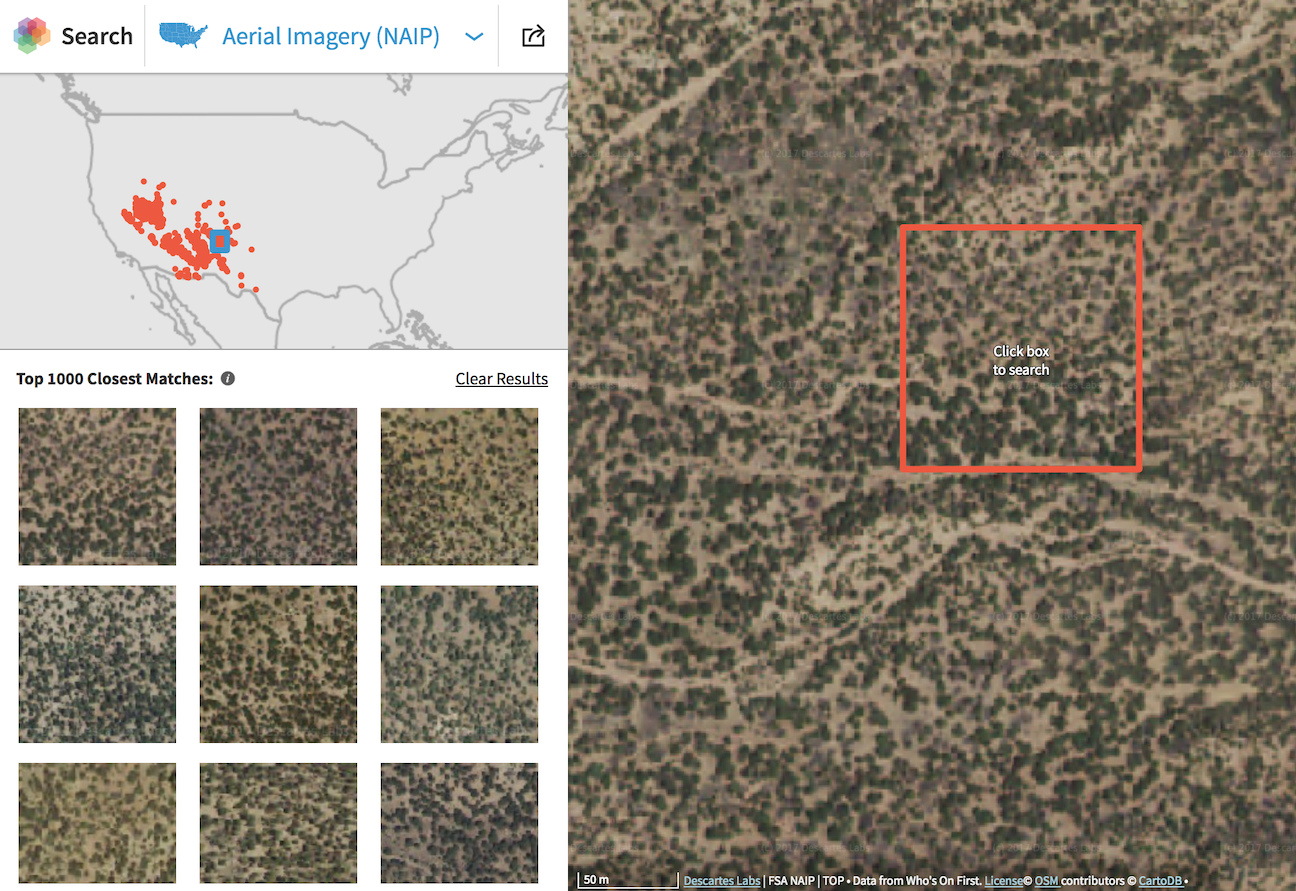}
\caption{The user interface shown here is publicly available at \url{https://search.descarteslabs.com}.  The query image and its local context are shown in the \textbf{right} panel.  The geographical distribution and thumbnail images of the search results are shown in the \textbf{top left} and \textbf{bottom left} panels, respectively.  In this particular search, the geographical extent of pi{\~n}on-juniper woodlands, common to the southwestern United States, is immediately visible.}
\label{fig:ui}
\end{figure}

\subsection{User Interface}
We built a browser-based user interface on top of the search back-end described above.  This interface was released publicly in March 2017 at \url{https://search.descarteslabs.com}.  We encourage the reader to try it out, but the basic structure is: a user clicks anywhere on the map, the top 1000 search results are displayed in thumbnail image form, and their geographical locations are shown on a map.

A screenshot of the user interface is shown in Figure~\ref{fig:ui}.  This particular search query highlights the benefit of showing the geographical distribution of the search results.  The user clicked on a single example of pi{\~n}on-juniper woodlands, a landscape common to the southwestern United States, and the full geographical extent of pi{\~n}on-juniper woodlands is visible immediately.

\section{Conclusion and Future Directions}
\label{sec:conclusion}
We have presented a system for visual search over billions of aerial and satellite images.  Binary features encoding visual information are extracted with a convolutional neural network that has been trained on aerial and satellite imagery.  Using a hash-based search method we are able to search over approximately 2 billion images in 0.1 seconds.  An interactive demo is available at \url{https://search.descarteslabs.com}

We see three clear future directions for extending the work presented here.
\begin{itemize}
    \item \textbf{Multi-scale search - }The current system searches at one spatial scale, namely square images that are 128 pixels across.  Instead we would like to enable search over a handful of spatial scales, both smaller and larger than 128 pixels across.
    \item \textbf{Geospatial filtering - }The current system searches across all images that have been indexed, regardless of their geographical location.  In the future we would like to enable geospatial filtering, e.g. only return results from Japan.
    \item \textbf{Temporal filtering - }The current system searches across a single image layer.  Although the images within this layer were acquired at different times, that temporal information has not been used in the search.  In the future we would like to enable temporal filtering, e.g. only return results that were acquired in May 2018.
\end{itemize}

With these changes, it should be possible, for example, to process and search over all Landsat 8 scenes, rather than searching over a single Landsat 8 composite image.  And while the work presented here used RGB imagery, the underlying technique should work with non-visual bands (infrared, SWIR, SAR, \textit{etc.}) or a different number of bands.  Such a system should provide a more flexible way to compactly encode visual information for large collections of aerial and satellite imagery.

\bibliographystyle{model2-names}
\bibliography{gvs1}

\begin{thebibliography}{21}
\expandafter\ifx\csname natexlab\endcsname\relax\def\natexlab#1{#1}\fi
\providecommand{\url}[1]{\texttt{#1}}
\providecommand{\href}[2]{#2}
\providecommand{\path}[1]{#1}
\providecommand{\DOIprefix}{doi:}
\providecommand{\ArXivprefix}{arXiv:}
\providecommand{\URLprefix}{URL: }
\providecommand{\Pubmedprefix}{pmid:}
\providecommand{\doi}[1]{\href{http://dx.doi.org/#1}{\path{#1}}}
\providecommand{\Pubmed}[1]{\href{pmid:#1}{\path{#1}}}
\providecommand{\bibinfo}[2]{#2}
\ifx\xfnm\relax \def\xfnm[#1]{\unskip,\space#1}\fi
\bibitem[{Babenko et~al.(2014)Babenko, Slesarev, Chigorin and
  Lempitsky}]{babenko2014neural}
\bibinfo{author}{Babenko, A.}, \bibinfo{author}{Slesarev, A.},
  \bibinfo{author}{Chigorin, A.}, \bibinfo{author}{Lempitsky, V.},
  \bibinfo{year}{2014}.
\newblock \bibinfo{title}{Neural codes for image retrieval}, in:
  \bibinfo{booktitle}{European conference on computer vision},
  \bibinfo{organization}{Springer}. pp. \bibinfo{pages}{584--599}.
\bibitem[{Datta et~al.(2008)Datta, Joshi, Li and Wang}]{datta2008image}
\bibinfo{author}{Datta, R.}, \bibinfo{author}{Joshi, D.}, \bibinfo{author}{Li,
  J.}, \bibinfo{author}{Wang, J.Z.}, \bibinfo{year}{2008}.
\newblock \bibinfo{title}{Image retrieval: Ideas, influences, and trends of the
  new age}.
\newblock \bibinfo{journal}{ACM Computing Surveys (Csur)} \bibinfo{volume}{40},
  \bibinfo{pages}{5}.
\bibitem[{Deng et~al.(2009)Deng, Dong, Socher, Li, Li and
  Fei-Fei}]{deng2009imagenet}
\bibinfo{author}{Deng, J.}, \bibinfo{author}{Dong, W.},
  \bibinfo{author}{Socher, R.}, \bibinfo{author}{Li, L.J.},
  \bibinfo{author}{Li, K.}, \bibinfo{author}{Fei-Fei, L.},
  \bibinfo{year}{2009}.
\newblock \bibinfo{title}{Imagenet: A large-scale hierarchical image database},
  in: \bibinfo{booktitle}{2009 IEEE conference on computer vision and pattern
  recognition}, \bibinfo{organization}{Ieee}. pp. \bibinfo{pages}{248--255}.
\bibitem[{Gudivada and Raghavan(1995)}]{gudivada1995content}
\bibinfo{author}{Gudivada, V.N.}, \bibinfo{author}{Raghavan, V.V.},
  \bibinfo{year}{1995}.
\newblock \bibinfo{title}{Content based image retrieval systems}.
\newblock \bibinfo{journal}{Computer} \bibinfo{volume}{28},
  \bibinfo{pages}{18--22}.
\bibitem[{He et~al.(2016)He, Zhang, Ren and Sun}]{he2016deep}
\bibinfo{author}{He, K.}, \bibinfo{author}{Zhang, X.}, \bibinfo{author}{Ren,
  S.}, \bibinfo{author}{Sun, J.}, \bibinfo{year}{2016}.
\newblock \bibinfo{title}{Deep residual learning for image recognition}, in:
  \bibinfo{booktitle}{Proceedings of the IEEE conference on computer vision and
  pattern recognition}, pp. \bibinfo{pages}{770--778}.
\bibitem[{Jing et~al.(2015)Jing, Liu, Kislyuk, Zhai, Xu, Donahue and
  Tavel}]{jing2015visual}
\bibinfo{author}{Jing, Y.}, \bibinfo{author}{Liu, D.},
  \bibinfo{author}{Kislyuk, D.}, \bibinfo{author}{Zhai, A.},
  \bibinfo{author}{Xu, J.}, \bibinfo{author}{Donahue, J.},
  \bibinfo{author}{Tavel, S.}, \bibinfo{year}{2015}.
\newblock \bibinfo{title}{Visual search at pinterest}, in:
  \bibinfo{booktitle}{Proceedings of the 21th ACM SIGKDD International
  Conference on Knowledge Discovery and Data Mining},
  \bibinfo{organization}{ACM}. pp. \bibinfo{pages}{1889--1898}.
\bibitem[{LeCun et~al.(1995)LeCun, Bengio et~al.}]{lecun1995convolutional}
\bibinfo{author}{LeCun, Y.}, \bibinfo{author}{Bengio, Y.}, et~al.,
  \bibinfo{year}{1995}.
\newblock \bibinfo{title}{Convolutional networks for images, speech, and time
  series}.
\newblock \bibinfo{journal}{The handbook of brain theory and neural networks}
  \bibinfo{volume}{3361}, \bibinfo{pages}{1995}.
\bibitem[{Liu et~al.(2007)Liu, Zhang, Lu and Ma}]{liu2007survey}
\bibinfo{author}{Liu, Y.}, \bibinfo{author}{Zhang, D.}, \bibinfo{author}{Lu,
  G.}, \bibinfo{author}{Ma, W.Y.}, \bibinfo{year}{2007}.
\newblock \bibinfo{title}{A survey of content-based image retrieval with
  high-level semantics}.
\newblock \bibinfo{journal}{Pattern recognition} \bibinfo{volume}{40},
  \bibinfo{pages}{262--282}.
\bibitem[{Lowe(2004)}]{lowe2004distinctive}
\bibinfo{author}{Lowe, D.G.}, \bibinfo{year}{2004}.
\newblock \bibinfo{title}{Distinctive image features from scale-invariant
  keypoints}.
\newblock \bibinfo{journal}{International journal of computer vision}
  \bibinfo{volume}{60}, \bibinfo{pages}{91--110}.
\bibitem[{Norouzi et~al.(2012)Norouzi, Fleet and
  Salakhutdinov}]{norouzi2012hamming}
\bibinfo{author}{Norouzi, M.}, \bibinfo{author}{Fleet, D.J.},
  \bibinfo{author}{Salakhutdinov, R.R.}, \bibinfo{year}{2012}.
\newblock \bibinfo{title}{Hamming distance metric learning}, in:
  \bibinfo{booktitle}{Advances in neural information processing systems}, pp.
  \bibinfo{pages}{1061--1069}.
\bibitem[{Rui et~al.(1999)Rui, Huang and Chang}]{rui1999image}
\bibinfo{author}{Rui, Y.}, \bibinfo{author}{Huang, T.S.},
  \bibinfo{author}{Chang, S.F.}, \bibinfo{year}{1999}.
\newblock \bibinfo{title}{Image retrieval: Current techniques, promising
  directions, and open issues}.
\newblock \bibinfo{journal}{Journal of visual communication and image
  representation} \bibinfo{volume}{10}, \bibinfo{pages}{39--62}.
\bibitem[{Salakhutdinov and Hinton(2009)}]{salakhutdinov2009semantic}
\bibinfo{author}{Salakhutdinov, R.}, \bibinfo{author}{Hinton, G.},
  \bibinfo{year}{2009}.
\newblock \bibinfo{title}{Semantic hashing}.
\newblock \bibinfo{journal}{International Journal of Approximate Reasoning}
  \bibinfo{volume}{50}, \bibinfo{pages}{969--978}.
\bibitem[{Sharif~Razavian et~al.(2014)Sharif~Razavian, Azizpour, Sullivan and
  Carlsson}]{sharif2014cnn}
\bibinfo{author}{Sharif~Razavian, A.}, \bibinfo{author}{Azizpour, H.},
  \bibinfo{author}{Sullivan, J.}, \bibinfo{author}{Carlsson, S.},
  \bibinfo{year}{2014}.
\newblock \bibinfo{title}{Cnn features off-the-shelf: an astounding baseline
  for recognition}, in: \bibinfo{booktitle}{Proceedings of the IEEE conference
  on computer vision and pattern recognition workshops}, pp.
  \bibinfo{pages}{806--813}.
\bibitem[{Smeulders et~al.(2000)Smeulders, Worring, Santini, Gupta and
  Jain}]{smeulders2000content}
\bibinfo{author}{Smeulders, A.W.}, \bibinfo{author}{Worring, M.},
  \bibinfo{author}{Santini, S.}, \bibinfo{author}{Gupta, A.},
  \bibinfo{author}{Jain, R.}, \bibinfo{year}{2000}.
\newblock \bibinfo{title}{Content-based image retrieval at the end of the early
  years}.
\newblock \bibinfo{journal}{IEEE Transactions on Pattern Analysis \& Machine
  Intelligence} , \bibinfo{pages}{1349--1380}.
\bibitem[{Tolias et~al.(2015)Tolias, Sicre and
  J{\'e}gou}]{tolias2015particular}
\bibinfo{author}{Tolias, G.}, \bibinfo{author}{Sicre, R.},
  \bibinfo{author}{J{\'e}gou, H.}, \bibinfo{year}{2015}.
\newblock \bibinfo{title}{Particular object retrieval with integral max-pooling
  of cnn activations}.
\newblock \bibinfo{journal}{arXiv preprint arXiv:1511.05879} .
\bibitem[{Vo and Hays(2019)}]{vo2019generalization}
\bibinfo{author}{Vo, N.}, \bibinfo{author}{Hays, J.}, \bibinfo{year}{2019}.
\newblock \bibinfo{title}{Generalization in metric learning: Should the
  embedding layer be embedding layer?}, in: \bibinfo{booktitle}{2019 IEEE
  Winter Conference on Applications of Computer Vision (WACV)},
  \bibinfo{organization}{IEEE}. pp. \bibinfo{pages}{589--598}.
\bibitem[{Warren et~al.(2016)Warren, Skillman, Chartrand, Kelton, Keisler,
  Raleigh and Turk}]{warren2016data}
\bibinfo{author}{Warren, M.S.}, \bibinfo{author}{Skillman, S.W.},
  \bibinfo{author}{Chartrand, R.}, \bibinfo{author}{Kelton, T.},
  \bibinfo{author}{Keisler, R.}, \bibinfo{author}{Raleigh, D.},
  \bibinfo{author}{Turk, M.}, \bibinfo{year}{2016}.
\newblock \bibinfo{title}{Data-intensive supercomputing in the cloud: Global
  analytics for satellite imagery}, in: \bibinfo{booktitle}{Data-Intensive
  Computing in the Clouds (DataCloud), 2016 Seventh International Workshop on},
  \bibinfo{organization}{IEEE}. pp. \bibinfo{pages}{24--31}.
\bibitem[{Yue-Hei~Ng et~al.(2015)Yue-Hei~Ng, Yang and
  Davis}]{yue2015exploiting}
\bibinfo{author}{Yue-Hei~Ng, J.}, \bibinfo{author}{Yang, F.},
  \bibinfo{author}{Davis, L.S.}, \bibinfo{year}{2015}.
\newblock \bibinfo{title}{Exploiting local features from deep networks for
  image retrieval}, in: \bibinfo{booktitle}{Proceedings of the IEEE conference
  on computer vision and pattern recognition workshops}, pp.
  \bibinfo{pages}{53--61}.
\bibitem[{Zhai et~al.(2017)Zhai, Kislyuk, Jing, Feng, Tzeng, Donahue, Du and
  Darrell}]{zhai2017visual}
\bibinfo{author}{Zhai, A.}, \bibinfo{author}{Kislyuk, D.},
  \bibinfo{author}{Jing, Y.}, \bibinfo{author}{Feng, M.},
  \bibinfo{author}{Tzeng, E.}, \bibinfo{author}{Donahue, J.},
  \bibinfo{author}{Du, Y.L.}, \bibinfo{author}{Darrell, T.},
  \bibinfo{year}{2017}.
\newblock \bibinfo{title}{Visual discovery at pinterest}, in:
  \bibinfo{booktitle}{Proceedings of the 26th International Conference on World
  Wide Web Companion}, \bibinfo{organization}{International World Wide Web
  Conferences Steering Committee}. pp. \bibinfo{pages}{515--524}.
\bibitem[{Zheng et~al.(2018)Zheng, Yang and Tian}]{zheng2018sift}
\bibinfo{author}{Zheng, L.}, \bibinfo{author}{Yang, Y.}, \bibinfo{author}{Tian,
  Q.}, \bibinfo{year}{2018}.
\newblock \bibinfo{title}{Sift meets cnn: A decade survey of instance
  retrieval}.
\newblock \bibinfo{journal}{IEEE transactions on pattern analysis and machine
  intelligence} \bibinfo{volume}{40}, \bibinfo{pages}{1224--1244}.
\bibitem[{Zheng et~al.(2016)Zheng, Zhao, Wang, Wang and Tian}]{zheng2016good}
\bibinfo{author}{Zheng, L.}, \bibinfo{author}{Zhao, Y.}, \bibinfo{author}{Wang,
  S.}, \bibinfo{author}{Wang, J.}, \bibinfo{author}{Tian, Q.},
  \bibinfo{year}{2016}.
\newblock \bibinfo{title}{Good practice in cnn feature transfer}.
\newblock \bibinfo{journal}{arXiv preprint arXiv:1604.00133} .

\end{thebibliography}

\end{document}